\documentclass{article}
\usepackage{ijcai19}

\usepackage{times}
\usepackage{soul}
\usepackage{url}
\usepackage[hidelinks]{hyperref}
\usepackage[utf8]{inputenc}
\usepackage[small]{caption}
\usepackage{graphicx}
\usepackage{amsmath}
\usepackage{booktabs}
\usepackage{algorithm}
\usepackage{algorithmic}
\usepackage{tikz}
\definecolor{light-blue}{rgb}{.6, .6, 1}
\definecolor{light-brown}{rgb}{.8, .4, .3}
\definecolor{light-yellow}{rgb}{1, .8, .5}
\definecolor{light-green}{rgb}{.7, 1, .7}

\newcommand\A{\mathcal{A}}
\newcommand\SSS{\mathcal{S}}
\newcommand\OO{\mathcal{O}}
\newcommand\M{\mathcal{M}}
\newcommand\R{\mathcal{R}}
\newcommand\IfO{\mbox{\emph{IfO}}}
\newcommand\IL{\mbox{\emph{IL}}}
\newcommand\RL{\mbox{\emph{RL}}}
\newcommand\BC{\mbox{\emph{BC}}}

\newcommand\IRL{\mbox{\emph{IRL}}}
\newcommand\GAIL{\mbox{\emph{GAIL}}}

\newcommand\TCN{\mbox{\emph{TCN}}}
\newcommand\BCO{\mbox{\emph{BCO}}}
\newcommand\GAN{\mbox{\emph{GAN}}}
\newcommand\MDP{\mbox{\emph{MDP}}}

\newcommand\ILPO{\mbox{\emph{ILPO}}}

\newcommand\LQR{\mbox{\emph{LQR}}}

\usepackage{fancyhdr}

\title{Recent Advances in Imitation Learning from Observation}

\author{
Faraz Torabi$^1$
\and
Garrett Warnell$^2$\And
Peter Stone$^1$
\affiliations
$^1$The University of Texas at Austin\\
$^2$Army Research Laboratory\\
\emails
\{faraztrb, pstone\}@cs.utexas.edu,
garrett.a.warnell.civ@mail.mil
}

\begin{document}
\thispagestyle{fancy}
\maketitle

\begin{abstract}
Imitation learning is the process by which one agent tries to learn how to perform a certain task using information generated by another, often more-expert agent performing that same task.
Conventionally, the imitator has access to both {\em state}  and {\em action} information generated by an expert performing the task (e.g., the expert may provide a kinesthetic demonstration of object placement using a robotic arm).
However, requiring the action information prevents imitation learning from a large number of existing valuable learning resources such as online videos of humans performing tasks.
To overcome this issue, the specific problem of \emph{imitation from observation} (\IfO) has recently garnered a great deal of attention, in which the imitator only has access to the {\em state} information (e.g., video frames) generated by the expert.
In this paper, we provide a literature review of methods developed for \IfO, and then point out some open research problems and potential future work.
\end{abstract}

\section{Introduction}
Imitation learning \cite{schaal1997learning,argall2009survey,osa2018algorithmic} is a problem in machine learning that autonomous agents face when attempting to learn tasks from another, more-expert agent.
The expert provides demonstrations of task execution, from which the imitator attempts to mimic the expert's behavior.
Conventionally, methods developed in this framework require the demonstration information to include not only the expert's {\em states} (e.g., robot joint angles), but also its {\em actions} (e.g., robot torque commands).
For instance, a human expert might provide a demonstration of an object-manipulation task to a robot by manually moving the robot's arms in the correct way, during which the robot can record both its joint angles and also the joint torques induced by the human demonstrator.
Unfortunately, requiring demonstration action information prevents imitating agents from using a large number of existing valuable demonstration resources such as online videos of humans performing a wide variety of tasks.
These resources provide state information (i.e., video frames) only---the actions executed by the demonstrator are not available.

In order to take advantage of these valuable resources, the more-specific problem of {\em imitation learning from observation} (\IfO) must be considered.
The \IfO ~problem arises when an autonomous agent attempts to learn how to perform tasks by observing {\em state-only} demonstrations generated by an expert.
Compared to the typical imitation learning paradigm described above, \IfO ~is a more natural way to consider learning from an expert, and exhibits more similarity with the way many biological agents appear to approach imitation.
For example, humans often learn how to do new tasks by observing other humans performing those tasks without ever having explicit knowledge of the exact low-level actions (e.g., muscle commands) that the demonstrator used.

Considering the problem of imitation learning using state-only demonstrations is not new \cite{ijspeert2002movement,bentivegna2002humanoid}.
However, with recent advances in deep learning and visual recognition, researchers now have much better tools than before with which to approach the problem, especially with respect to using raw visual observations.
These advances have resulted in a litany of new imitation from observation techniques in the literature, which can be categorized in several fundamentally-different ways.
In this paper, we offer an organization of recent \IfO ~research and then consider open research problems and potential future work.

\section{Background}
In this section, we first describe Markov decision processes (\MDP s), which constitute the foundation of all the algorithms presented in this paper.
We then provide background on conventional imitation learning, including the problem setting and a number of algorithms developed for that problem.

\subsection{Markov Decision Processes}
We consider artificial learning agents operating in the framework of Markov decision processes (\MDP s).
An \MDP ~can be described as a 6-tuple $\M = \{\SSS, \A, P, r, \gamma\}$, where $\SSS$ and $\A$ are state and action spaces, $P(s_{t+1}|s_t, a_t)$ is a function which represents the probability of an agent transitioning from state $s_t$ at time $t$ to $s_{t+1}$ at time $t+1$ by taking action $a_t$, $r:\SSS \times \A \rightarrow \R$ is a function that represents the reward feedback that the agent receives after taking a specific action at a given state, and $\gamma$ is a discount factor.
We denote by $o \in \OO$ visual observations, i.e., an image at time $t$ is denoted by $o_t$.
Typically, these visual observations can only provide partial state information.
$s$, on the other hand, constitutes the full proprioceptive state of the agent, and therefore is considered to provide complete state information.
In the context of the notation established above, the goal of reinforcement learning (\RL) \cite{sutton1998reinforcement} is to learn policies $\pi:\SSS \times \A \rightarrow [0,1]$---used to select actions at each state---that exhibit some notion of optimality with respect to the reward function.

\subsection{Imitation Learning}\label{IL}
In imitation learning (\IL), agents do not receive task reward feedback $r$.
Instead, they have access to expert demonstrations of the task and, from these demonstrations, attempt to learn policies $\pi$ that produce behaviors similar to that present in the demonstration.
Conventionally, these demonstrations are composed of the state and action sequences experienced by the demonstrator, i.e., expert demonstration trajectories are of the form $\tau_e = \{(s_t,a_t)\}$.
Broadly speaking, research in imitation learning area can be split into two main categories: {\em (1)} behavioral cloning (\BC), and {\em (2)} inverse reinforcement learning (\IRL).

\subsubsection{Behavioral Cloning}
Behavioral cloning \cite{bain1999framework,ross2011reduction,daftry2016learning} is a class of imitation learning algorithms where, given $\tau_e$, supervised learning is used to learn an imitating policy.
\BC ~has been used in a variety of the applications.
For instance, it has recently been used in the context of autonomous driving \cite{bojarski2016end} and in the context of autonomous control of aerial vehicles \cite{giusti2016machine}.
\BC ~is powerful in the sense that it requires only demonstration data to directly learn an imitation policy and does not require any further interaction between the agent and the environment.
However, \BC ~approaches can be rather brittle due to the well-known covariate shift problem \cite{ross2010efficient}.

\subsubsection{Inverse Reinforcement Learning}
The other major category of \IL ~approaches is composed of techniques based on inverse reinforcement learning \cite{russell1998learning,ng2000algorithms}.
Instead of directly learning a mapping from states to actions using the demonstration data, \IRL -based techniques iteratively alternate between using the demonstration to infer a hidden reward function and using \RL ~with the inferred reward function to learn an imitating policy.
\IRL -based techniques have been used for a variety of tasks such as maneuvering a helicopter \cite{abbeel2004apprenticeship} and object manipulation \cite{finn2016guided}.
Using \RL ~to optimize the policy given the inferred reward function requires the agent to interact with its environment, which can be costly from a time and safety perspective.
Moreover, the \IRL ~step typically requires the agent to solve an \MDP ~in the inner loop of iterative reward optimization \cite{abbeel2004apprenticeship,ziebart2008maximum}, which can be extremely costly from a computational perspective.
However, recently, a number of methods have been developed which do not make this requirement \cite{finn2016guided,ho2016generative,fu2018learning}.
One of these approaches is called generative adversarial imitation from observation (\GAIL) \cite{ho2016generative}, which uses an architecture similar to generative adversarial networks (\GAN s) \cite{goodfellow2014generative}, and the associated algorithm can be thought of as trying to induce an imitator state-action occupancy measure that is similar to that of the demonstrator.

\section{Imitation Learning from Observation}
We now turn to the problem that is the focus of this survey, i.e., that of imitation learning from observation (\IfO), in which the agent has access to state-only demonstrations (visual observations) of an expert performing a task, i.e., $\tau_e = \{o_t\}$.
As in \IL, the goal of the \IfO ~problem is to learn an imitation policy $\pi$ that results in the imitator exhibiting similar behavior to the expert.
Broadly speaking, there are two major components of the \IfO ~problem: {\em (1)} perception, and {\em (2)} control. 

\subsection{Perception}
Because \IfO ~depends on observations of a more expert agent, processing these observations perceptually is extremely important.
In the existing \IfO ~literature, multiple approaches have been used for this part of the problem.
One approach to the perception problem is to record the expert's movements using sensors placed directly on the expert agent \cite{ijspeert2001trajectory}.
Using this style of perception, previous work has studied techniques that can allow humanoid or anthropomorphic robots to mimic human motions, e.g., arm-reaching movements \cite{ijspeert2002movement,bentivegna2002humanoid}, biped locomotion \cite{nakanishi2004learning}, and human gestures \cite{calinon2007incremental}. 
A more recent approach is that of motion capture \cite{field2009motion}, which typically uses visual markers on the demonstrator to infer movement.
\IfO ~techniques built upon this approach have been used for a variety of tasks, including locomotion, acrobatics, and martial arts \cite{peng2018deepmimic,merel2017learning,setapen2010marionet}.
The methods discussed above often require costly instrumentation and pre-processing \cite{holden2016deep}.
Moreover, one of the goals of \IfO ~is to enable task imitation from available, passive resources such as YouTube videos, for which these methods are not helpful. 

Recently, however, convolutional neural networks and advances in visual recognition have provided promising tools to work towards visual imitation where the expert demonstration consists of raw video information (e.g., pixel color values).
Even with such tools, the imitating agent is still faced with a number of challenges: {\em (1)} embodiment mismatch, and {\em (2)} viewpoint difference.

\subsubsection{Embodiment Mismatch}
One challenge that might arise is if the demonstrating agent has a different embodiment from that of the imitator.
For example, the video could be of a human performing a task, but the goal may be to train a robot to do the same.
Since humans and robots do not look exactly alike (and may look quite different), the challenge is in how to interpret the visual information such that \IfO ~can be successful.
One \IfO ~method developed to address this problem learns a correspondence between the embodiments using autoencoders in a supervised fashion \cite{gupta2017learning}.
The autoencoder is trained in such a way that the encoded representations are invariant with respect to the embodiment features.
Another method learns the correspondence in an unsupervised fashion with a small amount of human supervision \cite{sermanet2018time}.

\subsubsection{Viewpoint Difference}
Another perceptual challenge that might arise in \IfO ~applications comes when demonstrations are not recorded in a controlled environment.
For instance, video background may be cluttered, or there may be mismatch in the point of view present in the demonstration video and that with which the agent sees itself.
One \IfO ~approach that attempts to address this issue learns a context translation model to translate an observation by predicting it in the target context \cite{liu2018imitation}.
The translation is learned using data that consists of images of the target context and the source context, and the task is to translate the frame from the source context to the that of the target.
Another approach uses a classifier to distinguish between the data that comes from different viewpoints and attempts to maximize the domain confusion in an adversarial setting during the training \cite{stadie2017third}.
Consequently, the extracted features can be invariant with respect to the viewpoint.

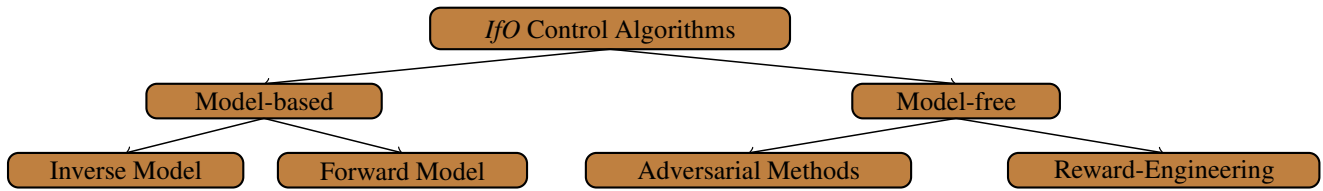
\begin{figure*}[!ht]
\centering
\begin{tikzpicture}[thick,scale=.92, every node/.style={scale=1.0}]

	\draw[use as bounding box, transparent] (-3,1.5) rectangle (16.6, 4);
	\draw[rounded corners,fill=brown,draw=black] (3.4,3.5) -- (3.4,4.1) -- (8.6,4.1) -- (8.6,3.5) -- cycle;
	\node [align=center] at (6, 3.75) {\IfO ~Control Algorithms};
	
	\draw[->, line width=0.2mm] (6,3.5) -- (1,3);
	\draw[rounded corners,fill=brown,draw=black] (-.7,2.5) -- (-.7,3) -- (2.7,3) -- (2.7,2.5) -- cycle;
	\node [align=center] at (1, 2.75) {Model-based};
	
	\draw[->, line width=0.2mm] (6,3.5) -- (11,3);
	\draw[rounded corners,fill=brown,draw=black] (9.5,2.5) -- (9.5,3) -- (12.5,3) -- (12.5,2.5) -- cycle;
	\node [align=center] at (11, 2.75) {Model-free};
	
	\draw[->, line width=0.2mm] (1,2.5) -- (-1,2);
	\draw[rounded corners,fill=brown,draw=black] (-2.7,1.5) -- (-2.7,2) -- (.7,2) -- (.7,1.5) -- cycle;
	\node [align=center] at (-1, 1.75) {Inverse Model};
	
	\draw[->, line width=0.2mm] (1,2.5) -- (3,2);
	\draw[rounded corners,fill=brown,draw=black] (1.2,1.5) -- (1.2,2) -- (4.8,2) -- (4.8,1.5) -- cycle;
	\node [align=center] at (3, 1.73) {Forward Model};
	
	\draw[->, line width=0.2mm] (11,2.5) -- (8,2);
	\draw[rounded corners,fill=brown,draw=black] (5.65,1.5) -- (5.65,2) -- (10.35,2) -- (10.35,1.5) -- cycle;
	\node [align=center] at (8, 1.75) {Adversarial Methods};
	
	\draw[->, line width=0.2mm] (11,2.5) -- (14,2);
	\draw[rounded corners,fill=brown,draw=black] (11.75,1.5) -- (11.75,2) -- (16.25,2) -- (16.25,1.5) -- cycle;
	\node [align=center] at (14, 1.73) {Reward-Engineering};

\end{tikzpicture}
\caption{A diagrammatic representation of categorization of the \IfO ~control algorithm. The algorithms can be categorized into two groups: (1) model-based algorithms in which the algorithms may use either a forward dynamics model \protect\cite{edwards2018imitating} or an inverse dynamics model \protect\cite{torabi2018behavioral,nair2017combining}. (2) Model-free algorithms, which itself can be categorized to adversarial methods \protect\cite{torabi2019generative,merel2017learning,stadie2017third} and reward engineering \protect\cite{sermanet2018time,gupta2017learning,liu2018imitation}.}
\label{fig:alg}
\end{figure*}

\subsection{Control}
Another main component of \IfO ~is control, i.e., the approach used to learn the imitation policy, typically under the assumption that the agent has access to clean state demonstration data $\{s_t\}$.
Since the action labels are not available, this is a very challenging problem, and many approaches have been discussed in the literature.
We organize \IfO ~control algorithms in the literature into two general groups: {\em (1)} model-based algorithms, and {\em (2)} model-free algorithms.
In the following, we discuss the features of each group and present relevant example  algorithms from the literature.

\subsubsection{Model-based}
Model-based approaches to \IfO ~are characterized by the fact that they learn some type of dynamics model during the imitation process.
The learned models themselves can be either {\em (1)} inverse dynamics models, or {\em (2)} forward dynamics model.

\paragraph{Inverse Dynamics Models} An inverse dynamics model is a mapping from state-transitions $\{(s_t, s_{t+1})\}$ to actions $\{a_t\}$ \cite{hanna2017grounded}.
One algorithm that learns and uses this kind of model for \IfO ~is that of \citeauthor{nair2017combining} \shortcite{nair2017combining}.
Given a single video demonstration, the goal of the proposed algorithm is to allow the imitator to reproduce the observed behavior directly.
To do so, the algorithm first allows the agent to interact with the environment using an exploratory policy to collect data $\{(s_t, a_t, s_{t+1})\}$.
Then, the collected data is used to learn a pixel-level inverse dynamics model which is a mapping from observation transition, $\{(o_t, o_{t+1})\}$, to actions, $\{a_t\}$.
Finally, the algorithm computes the actions for the imitator to take by applying the inverse dynamics model to the video demonstration.
Another algorithm of this type, reinforced inverse dynamics modeling \cite{torabi2019RIDM}, after learning the inverse dynamics model, uses a sparse reward function to further optimize the model. Then it executes the actions in the environment. It is shown that in most of the experiments the resulting behavior outperforms the expert.
Critically, these methods make the assumption that each observation transition is reachable through the application of a single action.
\citeauthor{pathak2018zeroshot} \shortcite{pathak2018zeroshot} attempt to remove this assumption by allowing the agent to execute multiple actions until it gets close enough to the next demonstrated frame.
Then this process is repeated for the next frame, and so on. 
All of the algorithms mentioned above attempt to {\em exactly} reproduce {\em single} demonstrations.
The authors \cite{torabi2018behavioral}, on the other hand, have proposed an algorithm, behavioral cloning from observation (\BCO), that is instead concerned with learning generalized imitation policies using multiple demonstrations.
The approach also learns an inverse dynamics model using an exploratory policy, and then uses that model to infer the actions from the demonstrations.
Then, however, since the states and actions of the demonstrator are available, a regular imitation learning algorithm (behavioral cloning) is used to learn the task. In another work, \citeauthor{guo2019hybrid} \shortcite{guo2019hybrid} proposed a hybrid algorithm that assumes that the agent also has access to both visual demonstrations and reward information as in the \RL ~problem.
A method similar to \BCO ~is formulated for imitating the demonstrations, and a gradient-based \RL ~approach is used to take advantage of the additional reward signal.
The final imitation policy is learned by minimizing a linear combination of the behavioral cloning loss and the \RL ~loss.


\paragraph{Forward Dynamics Models} A forward dynamics model is a mapping from state-action pairs, $\{(s_t, a_t)\}$, to the next states, $\{s_{t+1}\}$.
One \IfO ~approach that learns and uses this type of dynamics model is called imitating latent policies from observation (\ILPO) \cite{edwards2018imitating}.
\ILPO ~creates an initial hypothesis for the imitation policy by learning a latent policy $\pi(z|s_t)$ that estimates the probability of latent (unreal) action $z$ given the current state $s_t$.
Since actual actions are not needed, this process can be done offline without any interaction with the environment.
In order to learn the latent policy, they use a latent forward dynamics model which predicts $s_{t+1}$ and a prior over $z$ given $s_t$.
Then they use a limited number of environment interactions to learn an action-remapping network that associates the latent actions with their corresponding correct actions.
Since most of the process happens offline, the algorithm is efficient with regards to the number of interactions needed.

\subsubsection{Model-free}
The other broad category of \IfO ~control approaches is that of model-free algorithms.
Model-free techniques attempt to learn the imitation policy without any sort of model-learning step.
Within this category, there are two fundamentally-different types of algorithms: (1) adversarial methods, and (2) reward-engineering methods.


\paragraph{Adversarial Methods} Adversarial approaches to \IfO ~are inspired by the generative adversarial imitation learning (\GAIL) algorithm described in Section \ref{IL}.
Motivated by this work, \citeauthor{merel2017learning} \shortcite{merel2017learning} proposed an \IfO ~algorithm that assumes access to proprioceptive state-only demonstrations $\{s_t\}$ and uses a \GAN -like architecture in which the imitation policy is interpreted as the generator.
The imitation policy is executed in the environment to collect data, $\{(s_t^i, a_t^i)\}$, and single states are fed into the discriminator, which is trained to differentiate between the data that comes from the imitator and data that comes from the demonstrator.
The output value of the discriminator is then used as a reward to update the imitation policy using \RL .
Another algorithm, called \mbox{\emph{OptionGAN}} \cite{henderson2018optiongan}, uses the same algorithm combined with option learning to enable the agent to decompose policies for cases in which the demonstrations contain trajectories result from a diverse set of underlying reward functions rather than a single one.
In both of the algorithms discussed above, the underlying goal is to achieve imitation policies that generate a state distribution similar to the expert.
However, two agents having similar state distributions does not necessarily mean that they will exhibit similar behaviors.
For instance, in a ring-like environment, two agents that move with the same speed but different directions (i.e., one clockwise and one counter-clockwise) would result in each exhibiting the same state distribution even though their behaviors are opposite one another.
To resolve this issue, the authors \cite{torabi2019generative,torabi2019adversarial} proposed an algorithm similar to those above, but with the difference that the discriminator considers state {\em transitions}, $\{(s_t, s_{t+1})\}$, as the input instead of single states.
This paper also tested the proposed algorithm on the cases that the imitator has only access to visual demonstrations $\{o_t\}$, and showed that using multiple video frames instead of single frames resulted in good imitation policies for the demonstrated tasks. In this paper, the authors consider policies to be a mapping from observations $\{o_t\}$ to actions $\{a_t\}$. In follow up work \cite{torabi2019imitation}, motivated by the fact that agents often have access to their own internal states (i.e., \emph{proprioception}), proposed a modified version of this algorithm for the case of visual imitation that leverages this information in the policy learning process by considering a multi-layer perceptron (instead of convolutional neural networks) as the policy which maps internal states $s$ to actions $a$. Then it uses the observation $o$ as the input of the discriminator. By changing the architecture of the policy and leveraging the proprioceptive features, the authors showed that the performance improves significantly and the algorithm is much more sample efficient. In another follow up work \cite{torabi2019sample}, the authors modified the algorithm to make it more sample efficient in order to be able to execute it directly on physical robots. To do so, the algorithm was adapted in a way that linear quadratic regulators (\LQR) \cite{tassa2012synthesis} could be used for the policy training step. The algorithm is tested on a robotic arm which has resulted in reasonable performnce.
\citeauthor{zolna2018reinforced} \shortcite{zolna2018reinforced} has built on this work, and proposed an approach to adapt the algorithm to cases in which the imitator and the expert have different action spaces.
Instead of using consecutive states as the input of the discriminator, they use pairs of states with random time gaps, and show that this change helps improve imitation performance.
Another adversarial \IfO ~approach developed by \citeauthor{stadie2017third} \shortcite{stadie2017third} considers cases in which the imitator and demonstrator have different viewpoints.
To overcome this challenge, a new classifier is introduced that uses the output of early layers in the discriminator as input, and attempts to distinguish between the data coming from different viewpoints.
Then they train early layers of the discriminator and the classifier in such a way as to maximize the viewpoint confusion.
The intuition is to ensure that the early layers of the discriminator are invariant to viewpoint.
Finally, \citeauthor{sun2018provably} \shortcite{sun2018provably} have also developed an adversarial \IfO ~approach in which, from a given start state, a policy for each time-step of the horizon is learned by solving a minimax game.
The minimax game learns a policy that matches the state distribution of the next state given the policies of the previous time steps.


\paragraph{Reward Engineering} Another class of model-free approaches developed for \IfO ~control are those that utilize reward engineering.
Here, reward engineering means that, based on the expert demonstrations, a manually-designed reward function is used to find imitation policies via \RL.
Importantly, the designed reward functions are not necessarily the ones that the demonstrator used to produce the demonstrations---rather, they are simply estimates inferred from the demonstration data.
One such method, developed by \citeauthor{kimura2018internal} \shortcite{kimura2018internal}, first trains a predictor that predicts the demonstrator's next state given the current state.
The manually-designed reward function is then defined as the Euclidean distance of the actual next state and the one that the approximator returns.
An imitation policy is learned via \RL ~using the designed reward function.
Another reward-engineering approach is that of time-contrastive networks (\TCN) \cite{sermanet2018time}.
\TCN ~considers settings in which demonstrations are generated by human experts performing tasks and the agent is a robot with arms.
A triplet loss is used to train a neural network that is used to generate a task-specific state encoding at each time step.
This loss function brings states that occur in a small time-window closer together in the embedding space and pushes others farther apart.
The engineered reward function is then defined as the Euclidean distance between the embedded demonstration and the embedded agent's state at each time step, and an imitation policy is learned using \RL ~techniques.
\citeauthor{dwibedi2018learning} \shortcite{dwibedi2018learning} claims that, since \TCN ~uses single frames to learn the embedding function, it is difficult for \TCN ~to encode motion cues or the velocities of objects.
Therefore, they extend \TCN ~to the multi-frame setting by learning an embedding function that uses multiple frames as the input, and they show that it results in better imitation. 
Another approach of this type is developed by \citeauthor{goo2018learning} \shortcite{goo2018learning} in which the algorithm uses a formulation similar to shuffle-and-learn \cite{misra2016shuffle} to train a classifier that learns the order of frames in the demonstration.
The manually-specified reward function is then defined as the progress toward the task goal based on the learned classifier.
\citeauthor{aytar2018playing} \shortcite{aytar2018playing} also take a similar approach, learning an embedding function for the video frames based on the demonstration.
They use the closeness between the imitator's embedded states and some checkpoint embedded features as the reward function.
In another work, \citeauthor{gupta2017learning} \shortcite{gupta2017learning} consider settings in which the demonstrator and the imitator have different state spaces.
First, they train an autoencoder that maps states into an invariant feature space where corresponding states have the same features.
Then, they define the reward as the Euclidean distance of the expert and imitator state features in the invariant space at each time step.
Finally, they learn the imitation policy using this reward function with an \RL ~algorithm. 
\citeauthor{liu2018imitation} \shortcite{liu2018imitation} also uses the same reward function to solve the task however in a setting where the expert demonstrations and the imitator's viewpoints are different.

\section{Conclusion and Future Directions}
In this paper, we reviewed recent advances in imitation learning from observation (\IfO) and, for the first time, provided an organization of the research that is being conducted in this field.
In this section, we provide some directions for future work.

\subsection{Perception}
Adversarial training techniques have led to several recent and exciting advances in the computer vision community.
One such advance is in the area of pose estimation \cite{cao2017realtime,wang20193d}, which enables detection of the position and orientation of the objects in a cluttered video through keypoint detection---such keypoint information may also prove useful in \IfO.
While there has been a small amount of effort to incorporate these advances in \IfO ~\cite{peng2018sfv}, there is still much to investigate.

Another recent advancement in computer vision is in the area of visual domain adaptation \cite{wang2018deep}, which is concerned with transferring learned knowledge to different visual contexts.
For instance, the recent success of CycleGAN \cite{zhu2017unpaired} suggests that modified adversarial techniques may be applicable to \IfO ~problems that require solutions to embodiment mismatch, though it remains to be seen if such approaches will truly lead to advances in \IfO.

\subsection{Application on Physical Robots}
Very few of the \IfO ~algorithms discussed have actually been successfully tested on physical robots, such as \cite{sermanet2018time,liu2018imitation}.
That is, most discuss results only in simulated domains.
For instance, while adversarial methods currently provide state-of-the-art performance for a number of baseline experimental \IfO ~problems, these methods exhibit high sample complexity and have therefore only been applied to relatively simple simulation tasks.
Thus, an open problem in \IfO ~is that of finding ways to adapt these techniques such that they can be used in scenarios for which high sample complexity is prohibitive, i.e., tasks in robotics.

\subsection{Integration}
The papers reviewed in this survey are exclusively concerned with the \IfO ~problem, i.e., finding imitation policies from state-only demonstrations.
However, to achieve the overall goal of developing fully-intelligent agents, algorithms that have been developed for other learning paradigms (e.g., \RL) should be integrated with these techniques.
While there is some previous work that considers a combination of imitation learning and \RL ~\cite{zhu2018reinforcement} or \IfO ~and \RL ~\cite{guo2019hybrid}, there is still much to investigate.

\section*{Acknowledgments}
This work has taken place in the Learning Agents Research
Group (LARG) at the Artificial Intelligence Laboratory, The University
of Texas at Austin.  LARG research is supported in part by grants from
the National Science Foundation (IIS-1637736, IIS-1651089,
IIS-1724157), the Office of Naval Research (N00014-18-2243), Future of
Life Institute (RFP2-000), Army Research Lab, DARPA, Intel, Raytheon,
and Lockheed Martin.  Peter Stone serves on the Board of Directors of
Cogitai, Inc.  The terms of this arrangement have been reviewed and
approved by the University of Texas at Austin in accordance with its
policy on objectivity in research.

\bibliographystyle{named}
\bibliography{ijcai19}

\end{document}